%% file: sample-sigconf.tex
\documentclass[sigconf]{acmart}
\usepackage{booktabs} 
\usepackage{graphicx}
\usepackage{subfigure}

\copyrightyear{2018} 
\acmYear{2018} 
\setcopyright{acmcopyright}
\acmConference[CSCS 2018]{2nd Computer Science in Cars Symposium - Future Challenges in Artificial Intelligence & Security for Autonomous Vehicles}{September 13--14, 2018}{Munich, Germany}
\acmBooktitle{2nd Computer Science in Cars Symposium - Future Challenges in Artificial Intelligence & Security for Autonomous Vehicles (CSCS 2018), September 13--14, 2018, Munich, Germany}
\acmPrice{15.00}
\acmDOI{10.1145/3273946.3273947}
\acmISBN{978-1-4503-6616-8/18/09}

\acmConference[CSCS'18]{ACM COMPUTER SCIENCE IN CARS SYMPOSIUM}{September 2018}{Munich, Germany}

\acmArticle{4}

\editor{Jennifer B. Sartor}
\editor{Theo D'Hondt}
\editor{Wolfgang De Meuter}

\begin{document}

\title[\resizebox{4in}{!}{AutoRVO: Local Navigation with Dynamic Constraints in Dense Heterogeneous Traffic}]{AutoRVO: Local Navigation with Dynamic Constraints in Dense Heterogeneous Traffic}

\author{Yuexin Ma}
\affiliation{
  \institution{The University of Hong Kong}}
\email{yxma@cs.hku.hk}

\author{Dinesh Manocha}
\affiliation{
  \institution{University of Maryland at College Park}}
\email{dm@cs.umd.edu}
\email{http://gamma.cs.unc.edu/CTMAT}

\author{Wenping Wang}
\affiliation{
  \institution{The University of Hong Kong}}
\email{wenping@cs.hku.hk}

\begin{abstract}
We present a novel algorithm for simulating heterogeneous road-agents such as cars, tricycles, bicycles, and pedestrians in dense traffic by computing collision-free navigation. Our approach computes smooth trajectories for each agent by taking into account the dynamic constraints. We describe an efficient optimization-based algorithm for each road-agent based on reciprocal velocity obstacles that takes into account kinematic and dynamic constraints. Our algorithm uses tight fitting shape representations based on medial axis to compute collision-free trajectories in dense traffic situations. We evaluate the performance of our simulation algorithm in real-world dense traffic scenarios and highlight the benefits over prior reciprocal collision avoidance schemes.  
\end{abstract}



\maketitle

\input{samplebody-conf}

\bibliographystyle{ACM-Reference-Format}
\bibliography{sample-bibliography}

\end{document}

%% file: samplebody-conf.tex
\section{Introduction}

\begin{figure}[!t]
\centering
\subfigure[]{
\label{fig:trafficView}
\includegraphics[width=0.97\columnwidth]{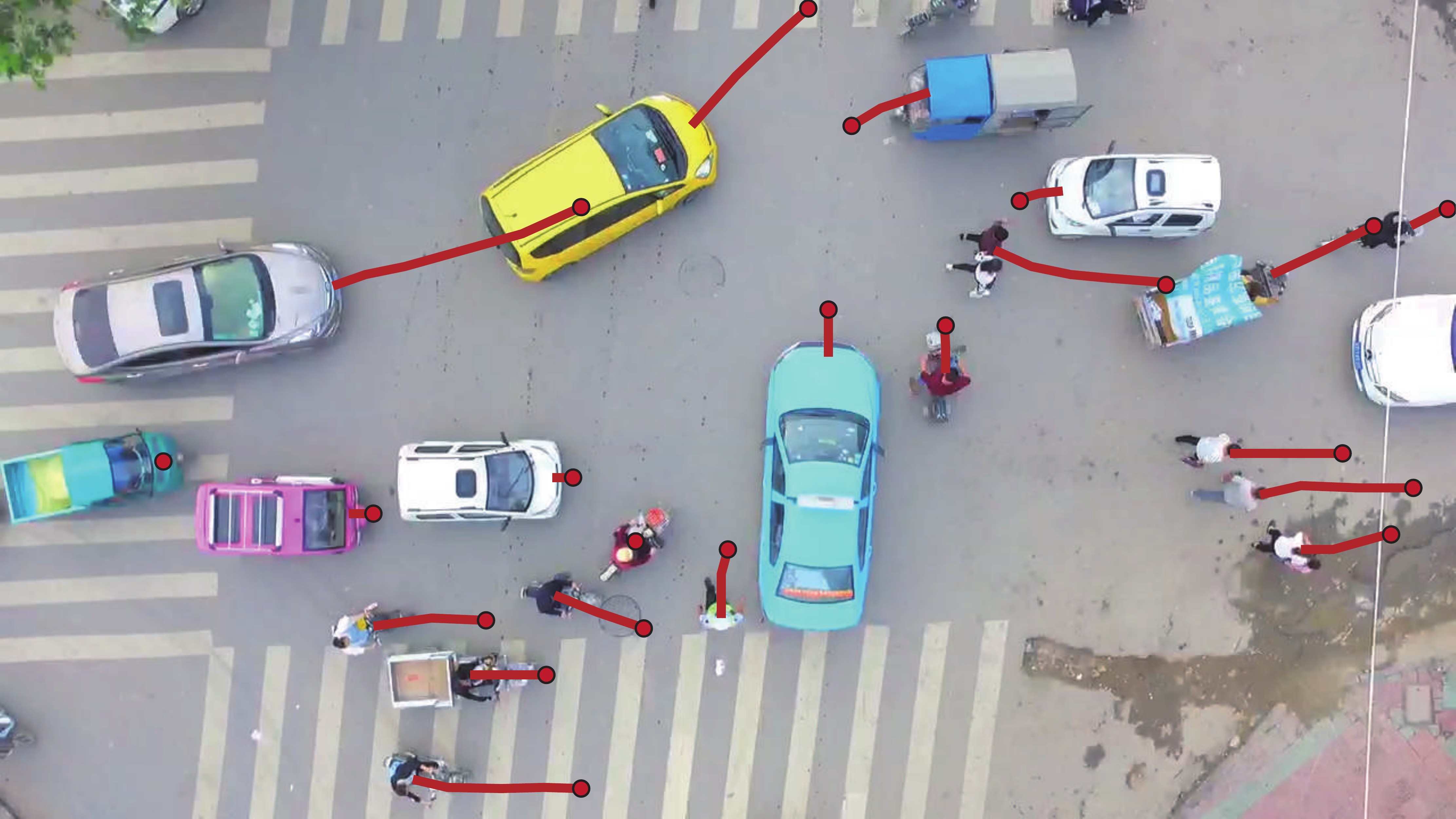}}
\centering
\subfigure[]{
\label{fig:trafficView_sim}
\includegraphics[width=0.975\columnwidth]{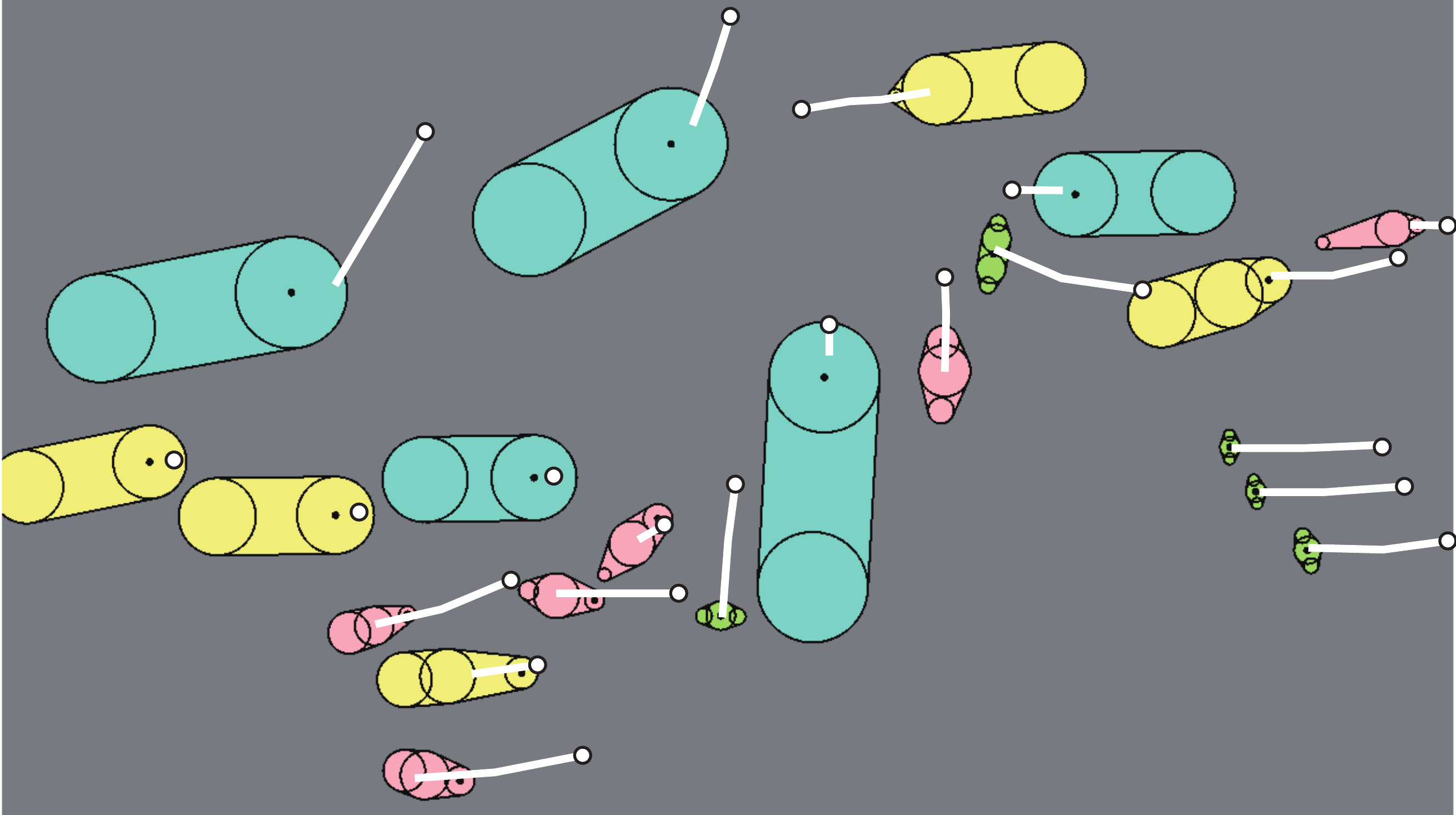}}
\caption{\textbf{Dense traffic and navigation}: (a) One frame of a top-down-view dense traffic video with different vehicles and pedestrians from a real-world scene. We highlight different road-agents in this dense scenario. Red curves show trajectories of road-agents in 50 frames of the video; (b) Simulated traffic scenario with our medial-axis-based agent representation agent representation. We model heterogeneous road-agents corresponding to pedestrians, bicycles, tricycles, and cars, as green, pink, yellow, and blue, respectively. Our algorithm, AutoRVO, models the dynamics of these agents and computes collision-free trajectories in similar time of (a) shown as white curves. We observe high accuracy with real-world vehicle and pedestrian trajectories.}
\label{fig:traffic}
\vspace{-3ex}
\end{figure}

Multi-agent local navigation is an important problem in robotics, crowd simulation, traffic simulation, and traffic modeling. At a broad level, the goal is to compute a collision-free trajectory for each road-agent in a distributed manner. Furthermore, it is important to satisfy other constraints corresponding to kinematics, dynamics, and smoothness. Some of the most widely used algorithms for decentralized multi-agent navigation are based on velocity obstacles~\cite{fiorini1998motion}. They have been extended to perform reciprocal collision avoidance between a large number of active agents and applied to simulate human-like crowds~\cite{van2011reciprocal} and multiple car-like robots~\cite{alonso2012reciprocal}. Furthermore, they have been extended to take dynamic constraints into account~\cite{wilkie2009generalized,bareiss2015generalized}.

There is considerable interest in developing multi-agent navigation algorithms for autonomous driving and simulating real-world traffic scenarios~\cite{cheung2018identifying,ziegler2014making,turri2013linear,best2017autonovi, luo2018autonomous, cheung2018identifying}. These algorithms consider dynamic constraints of vehicles, environmental factors, and traffic rules. However, current autonomous driving navigation algorithms are limited to simple scenarios with sparse traffic or few vehicles that are moving in specified lanes and simple traffic conditions. They do not model the movement of pedestrians or bicycles or their close interactions with vehicles or two-wheelers (i.e. road-agents), and instead maintain large distances for safety. As a result, current autonomous driving simulators or systems cannot model dense traffic scenarios with heterogeneous road-agents of varying shapes and dynamics. 

\begin{figure}[!t]
\centering
\subfigure[]{
\label{fig:represent1}
\includegraphics[width=0.48\columnwidth]{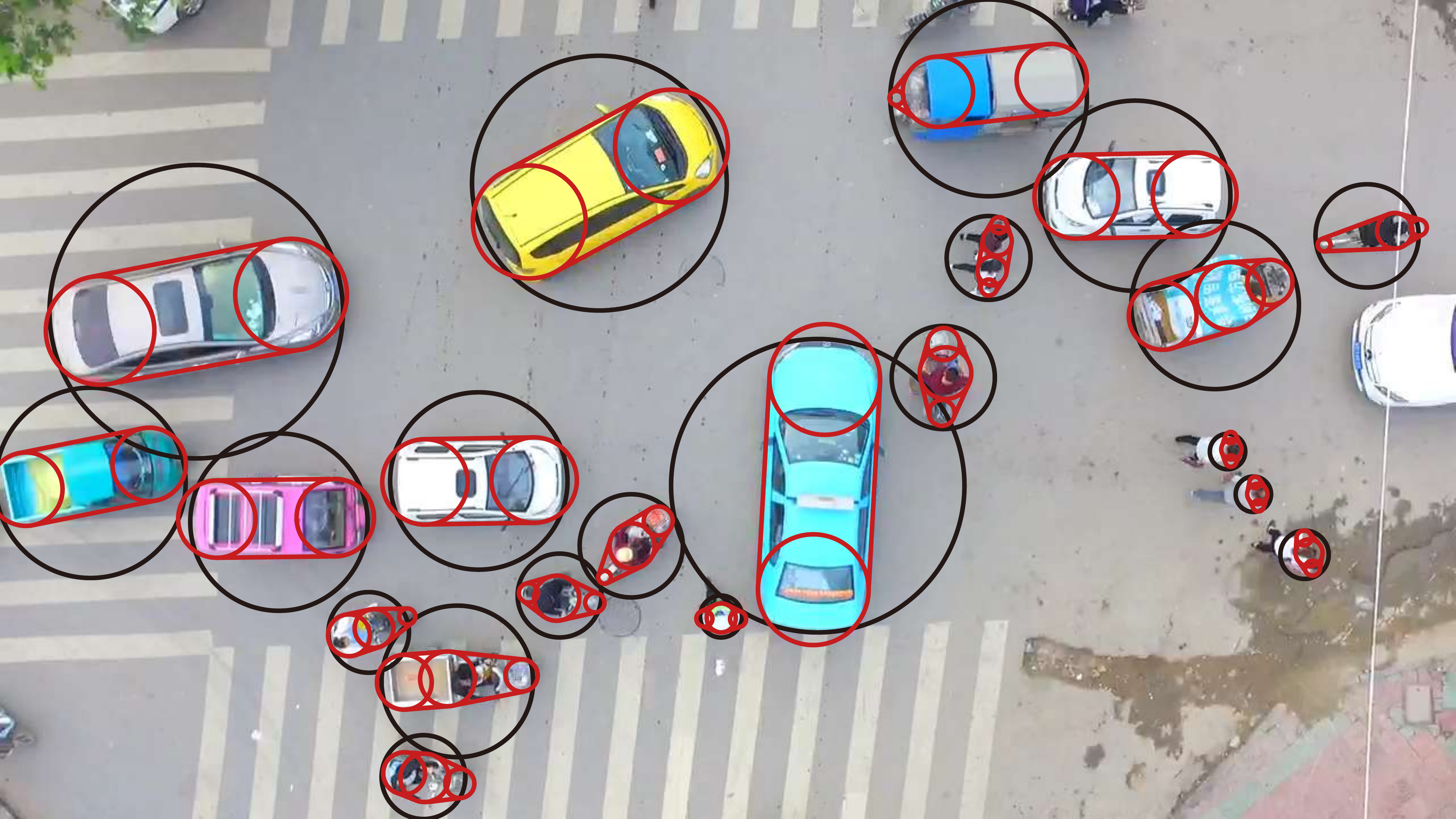}}
\centering
\subfigure[]{
\label{fig:represent2}
\includegraphics[width=0.48\columnwidth]{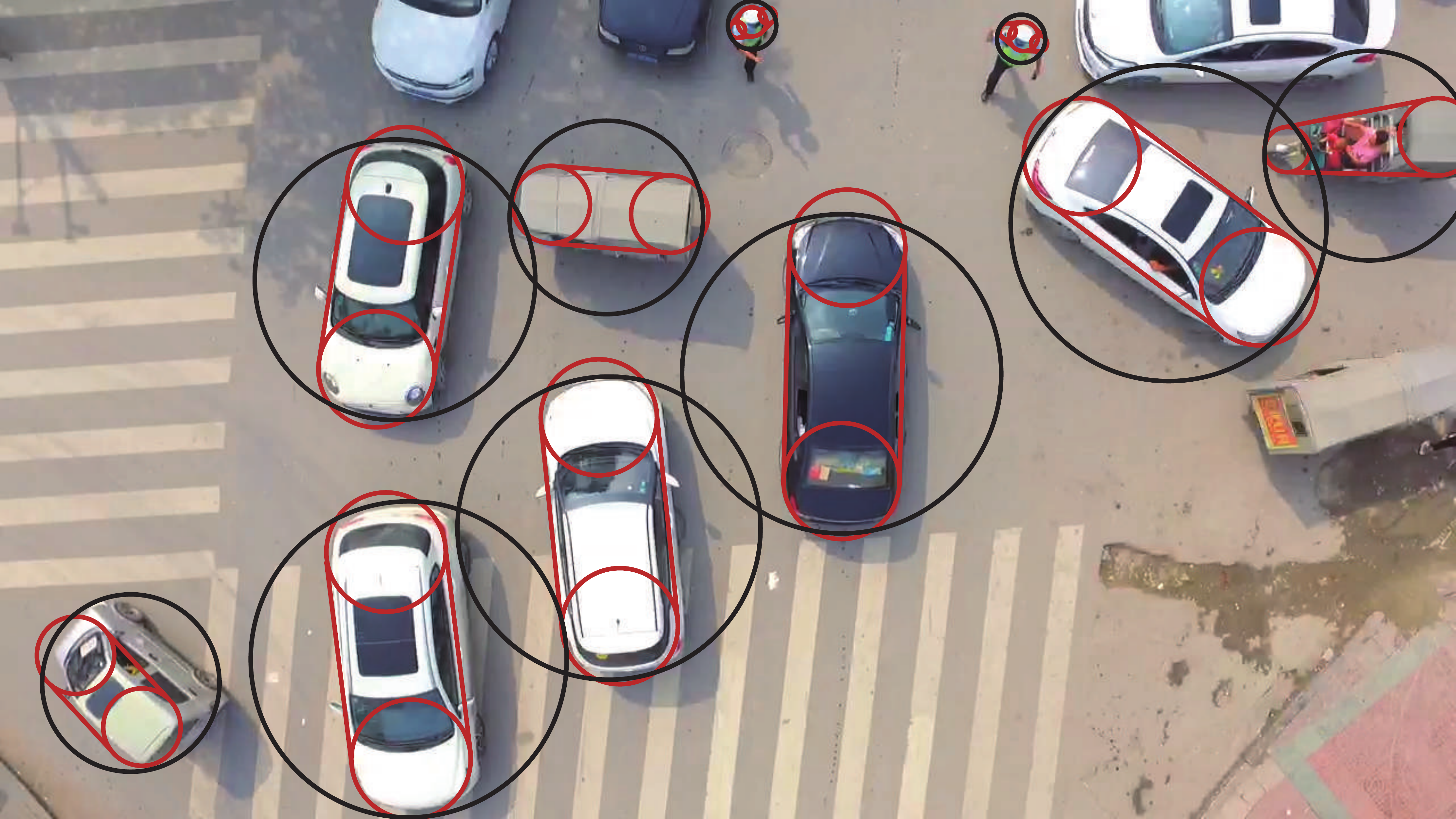}}
\caption{Comparison between the representation of disk (black) and our medial-axis-based representation (CTMAT in red). Our representation is less conservative and can accurately model such dense traffic scenarios. On the other hand, disk-based representations are overly conservative and unable to compute collision free trajectories. }
\label{fig:representation}
\vspace{-3ex}
\end{figure}

Most prior work on reciprocal collision avoidance is limited to simple agent shapes, including circles or ellipses. They use geometric properties of these simple shapes to design efficient navigation algorithms. However, such disk-based agent representations can be overly conservative in dense traffic scenarios which may consist of large or small vehicles, pedestrians, bicycles, two-wheelers, etc. in close proximity to each other(Fig.~\ref{fig:representation}). As a result, we need more efficient and less conservative multi-agent navigation algorithms that can simulate real-world traffic scenarios.

There is also considerable recent interest in developing simulators to evaluate the performance of autonomous driving algorithms~\cite{best2018autonovi}. The main goal is to generate traffic scenarios with different number of entities, driver behaviors and road conditions, that can be used to predict the performance of the sensors and navigation strategies of the autonomous vehicle in those scenarios. It allows rapid development and testing of vehicle configurations and can reduce labor costs and strengthen safety.

\textbf{Main Contributions:} We present a novel multi-agent simulation algorithm, AutoRVO, for local navigation with dynamic constraints in dense, heterogeneous scenarios. Our approach is designed for traffic-like environments, which have different agents of varying shapes and different dynamic constraints. Our approach assumes that the exact positions and velocities of all the agents are known. Our algorithm uses the CTMAT-based agent representation~\cite{ma2018efficient} and computes a new velocity for each agent based on local optimization. A key aspect of our approach is that it can handle the non-linear dynamics of different vehicles and compute collision-free trajectories in dense situations without making any assumptions about their movement patterns or trajectories. We have evaluated the performance of our algorithm on real-world traffic scenarios with vehicles that are different in terms of size and dynamic constraints, pedestrians, and bicycles. We computed their trajectories using AutoRVO and compared them with the real-world trajectories extracted from videos by using a tracking algorithm~\cite{xiang2015learning} . We evaluate the accuracy based on the Entropy metric~\cite{guy2012statistical} and observe considerable improvement over prior multi-agent navigation schemes. Overall, AutoRVO is the first algorithm that can generate collision-free trajectories for road-agents in dense scenarios. Furthermore, it can be used to generate plausible traffic behaviors and configurations for autonomous driving simulators. 

The rest of the paper is organized as follows. Section 2 offers an overview of related work in local navigation with dynamic constraints. We introduce the kinematic model and state space of vehicles and road-agents in Section 3. In Section 4, we give a detailed description of our novel multi-agent navigation algorithm, AutoRVO. We highlight the performance of our algorithm on challenging scenarios in Section 5 and analyze algorithm complexity in Section 6.

\section{Related Work}
In this section, we give a brief overview of prior work on collision avoidance, motion planning,  kinematic and dynamic modeling, and autonomous navigation. 

\subsection{Multi-Agent Navigation using Velocity Obstacles}
Based on the approach of velocity obstacles (VO)~\cite{fiorini1998motion}, reciprocal collision avoidance ORCA~\cite{van2011reciprocal} allows each agent to take half of the responsibility for avoiding pairwise collisions. In the ORCA algorithm, each agent has the same state space and can change its velocity instantaneously. Some subsequent approaches take various dynamic constraints into consideration, such as AVO algorithm~\cite{van2011reciprocal}, which computes free and collision-free velocities by using velocity-space reasoning with acceleration constraints; CCO algorithm~\cite{rufli2013reciprocal}, which considers continuous control obstacles; and LQR-obstacles algorithm~\cite{bareiss2013reciprocal}. All these techniques are designed for circular agents. The ORCA algorithm has been extended to elliptical agents~\cite{best2016real}. However, they are not applicable for arbitrary-shaped agents.

\subsection{Traffic and Autonomous Driving Simulators}
There are many works on simulating traffic for research or commercial purposes. Some methods focus on simulating huge traffic flow: ~\cite{horni2016multi,sewall2011interactive} simulate large traffic networks of thousands of vehicles, SUMO~\cite{krajzewicz2002sumo} provides a modular agent-based simulation network from a 2D perspective. However, it is not clear whether these methods can generate the trajectories or behaviors of different road-entities corresponding to vehicles, buses, two-wheelers, bicycles, pedestrians, in dense or congested scenarios. There are already some widely used simulators, like TORCS~\cite{wymann2000torcs} and CARLA~\cite{dosovitskiy2017carla}. But they lack of dense scenarios with heterogeneous vehicles and pedestrians. Recently, work on simulating traffic has increased the fidelity of modelled drivers and vehicles~\cite{chen2010review}. AutonoVi-Sim~\cite{best2018autonovi} is a high-fidelity simulation platform, which supports multiple vehicles with steering and acceleration limits and overall vehicle dynamic profiles. Many studies of autonomous driving~\cite{pendleton2017perception,katrakazas2015real,saifuzzaman2014incorporating,anderson2011design,likhachev2009planning} use such simulators as a tool to show their performance. However, the common limitation of these works is that they are designed for simple scenarios and could not be directly used in dense traffic scenarios with different kinds of vehicles and pedestrians.

\subsection{Kinematic and Dynamic Modeling}
There are many approaches to model vehicles with kinematic and dynamic constraints. Some of the simplest methods are based on the linear dynamics of vehicles~\cite{lavalle2006planning}, but these may not be accurate. Other methods make use of non-linear dynamic forces~\cite{borrelli2005mpc}, which are more accurate, but the resulting algorithms are more time consuming. The Reeds-Shepp formulation~\cite{reeds1990optimal} supports the forward and backward motion of a car. Other kinematic and dynamic models for a moving car are described in~\cite{margolis1991multipurpose}. 

Many velocity-obstacle-based methods have been extended to account for dynamic constraints, including  differential-drive~\cite{alonso2013optimal}, double-integrator~\cite{lalish2012distributed}, arbitrary integrator~\cite{rufli2013reciprocal}, car-like~\cite{alonso2012reciprocal}, linear quadratic regulator (LQR) controllers~\cite{bareiss2013reciprocal}, non-linear equations of motion~\cite{bareiss2015generalized}, etc. Some other algorithms, like NH-ORCA~\cite{alonso2013optimal}, transfer non-linear equations of motion into a linear formulation. Based on non-linear velocity obstacles NLVO~\cite{shiller2001motion}, and GVO~\cite{wilkie2009generalized}, GRVO~\cite{bareiss2015generalized} can account for non-homogeneous agents with nonlinear equations of motion. However, all these methods are restricted to simple disc-based representations and can be overly conservative in terms of handling dense scenarios and heterogeneous agents. 

\subsection{Maneuver Planning for Autonomous Driving}
There is considerable work on maneuver planning of autonomous vehicles, including driving corridors ~\cite{hardy2013contingency}, potential-field methods~\cite{galceran2015toward}, random-exploration ~\cite{kuwata2009real}, occupancy grids methods~\cite{kolski2006autonomous}, etc. Some approaches limit the vehicles to staying in lanes to avoid collisions with obstacles or other vehicles~\cite{turri2013linear,fritz2004chauffeur} or consider a driver's behaviors~\cite{sadigh2016planning,galceran2015multipolicy,best2017autonovi}. They are not applicable for complex traffic without clear lanes and signals.

\section{Problem Formulation and Notation}
In this section, we introduce the notation, representation of road-agents, kinematic and dynamic models of different vehicles, and their state space.

\subsection{Representation and Kinematic Models}

\begin{figure}[!t]
\centering
\includegraphics[width=0.99\columnwidth]{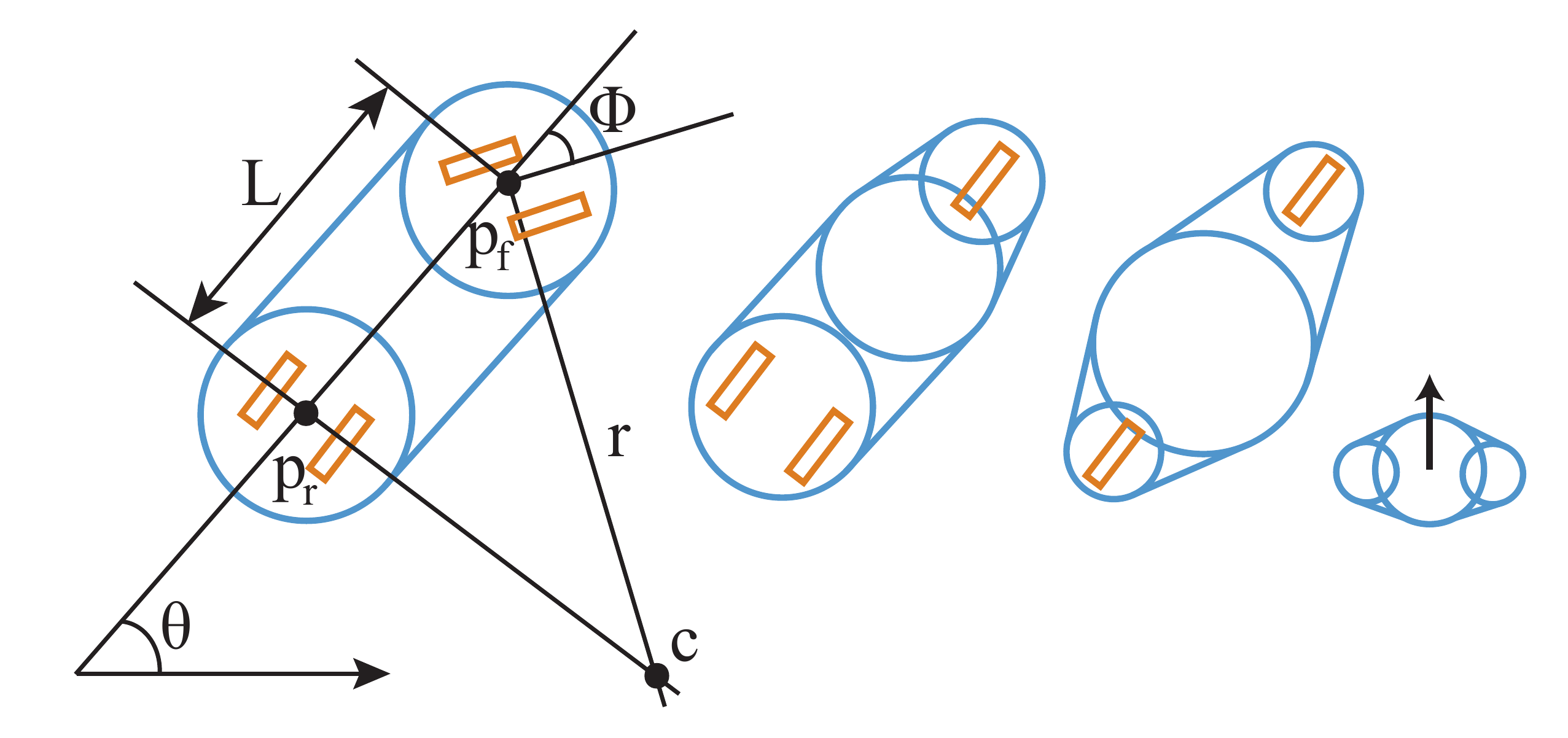}
\caption{\textbf{Representation of vehicles and pedestrians}. From left to right are the agent representations for a car, a tricycle, a bicycle, and a pedestrian. The blue shape is our CTMAT agent representation, the brown rectangles denote the tires, and the black arrow on the pedestrian indicates the forward-facing direction.}
\label{fig:model}
\vspace{-3ex}
\end{figure}

To represent different shapes corresponding to  heterogeneous vehicles in the real world, we use the  CTMAT representation~\cite{ma2018efficient}. CTMAT is a medial-axis-based representation that can provide a tighter fitting shape for different road-agents.  The underlying representation consists of circles and tangent line segments between any pair of adjacent circles. Fig.~\ref{fig:model} shows four examples of CTMAT for different vehicles and pedestrians. A key issue is to model the dynamic constraints of different agents. Therefore, we extend the simple-car kinematic model~\cite{lavalle2006planning,laumond1998guidelines} to different vehicle or agent types. As the figure indicates, if the steering wheel is turned, the vehicle will rotate around the center $c$, which is determined by the steering angle $\phi$ and body length $L$. Let's assume that the orientation is represented as $\theta$ and the speed is denoted as $v$. The vehicle's motion can be denoted as follows:

 \begin{equation*}
   \dot{\vec p} = (v\cos(\theta), v\sin(\theta)),  \qquad  \dot{\theta} = \frac{\tan(\phi)}{L}v.
  \end{equation*}

\noindent We give more details on computing $ \dot{v}$ and $\dot{\phi}$ in Section 4. Pedestrians do not exhibit steering behaviors as their dynamic constraints are different from that of vehicles. Instead, they can change their orientation instantly and always move according to their forward-facing direction.

\subsection{State Space}
The simulator state includes all the entities in the scenario, including all obstacles and agents. We always use an n-dimensional space to describe an agent's physical state and properties. Our approach is designed for different road-agents corresponding to pedestrians, bicycles, tricycles, and cars with different shapes. The state space of the pedestrians is denoted as $X_p = \{ T, v, v^o, \theta, \theta^o\}$. $T$ records the components of CTMAT representation, including circles and their tangent line segments. $v$ and $\theta$ denote current speed and orientation, respectively. $v^o$ and $\theta^o$ are the preferred speed and the preferred orientation, respectively.

The bicycles, tricycles, and cars have kinematic and dynamic constraints on their turning motion. The state space for vehicles is represented as $X_v = \{ T, p_f, p_r, v, \phi, v^o, \phi^o, u_t, u_\phi, \theta, b\}$ (see Fig.~\ref{fig:model}). $T$ also stands for the CTMAT representation like that of pedestrians. $p_f$ is the position of the front wheel for bicycles and tricycles, or the position of the middle point between two front wheels for cars. $p_r$ is similar to $p_f$ but for rear wheels. $v$ and $\phi$ represent vehicles' speed and steering, respectively. $v^o$ and $\phi^o$ are preferred speed and steering, respectively. Every vehicle has two degrees of control, throttle $u_t$ and steering $u_\phi$. We define $-1\leq u_t \leq 1$, where $-1$ denotes the maximum braking effort and $1$ represents the maximum throttle. $-1\leq u_\phi \leq 1$ indicates the steering effort from $-\phi_{max}$ to $\phi_{max}$. The boundary values of these dynamic variables are distinctive for different types of vehicles. $\theta$ stands for the orientation. $b$ is a label to record a vehicle's current behavior (turn left or turn right, wait, or go ahead), which is related to the choice of dynamic constraints. 
 
In addition, we define a label $C_{type}$ to record the road-agent's type, i.e. $1$ for pedestrian, $2$ for bicycle, so that they are distinguishable from each other. We regard the middle point of $T$ as the reference point of pedestrians and $p_f$ as the reference point of vehicles. 

\section{AutoRVO: Our Navigation Algorithm}
We assume that each road-agent has smart sensors that can capture surrounding environmental information such as nearby obstacles and the current speed, steering, position, and orientation of other agents. Our approach is based on a reciprocal collision avoidance method and uses optimization method to compute a local trajectory.  In particular, our algorithm proceeds using three main steps: first, we compute the preferred speed and steering $(v^o, \phi^o)$ for vehicles and preferred speed and orientation $(v^o, \theta^o)$ for pedestrians. Second, we sample around $(v^o, \phi^o)$ or $(v^o, \theta^o)$ to get a set of solution candidates for new speed and steering or orientation. Finally, we use an optimization function (Equation.6) to select the best solution for $(v, \phi)$ or $(v, \theta)$. After that, we update the state of each road-agent and change its trajectory for time $\tau$. For each step, we present the details for the vehicles, and then explain the differences for pedestrians.

\begin{figure}[!t]
\subfigure[]{
\label{fig:searchSpace_1}
\includegraphics[width=0.48\columnwidth]{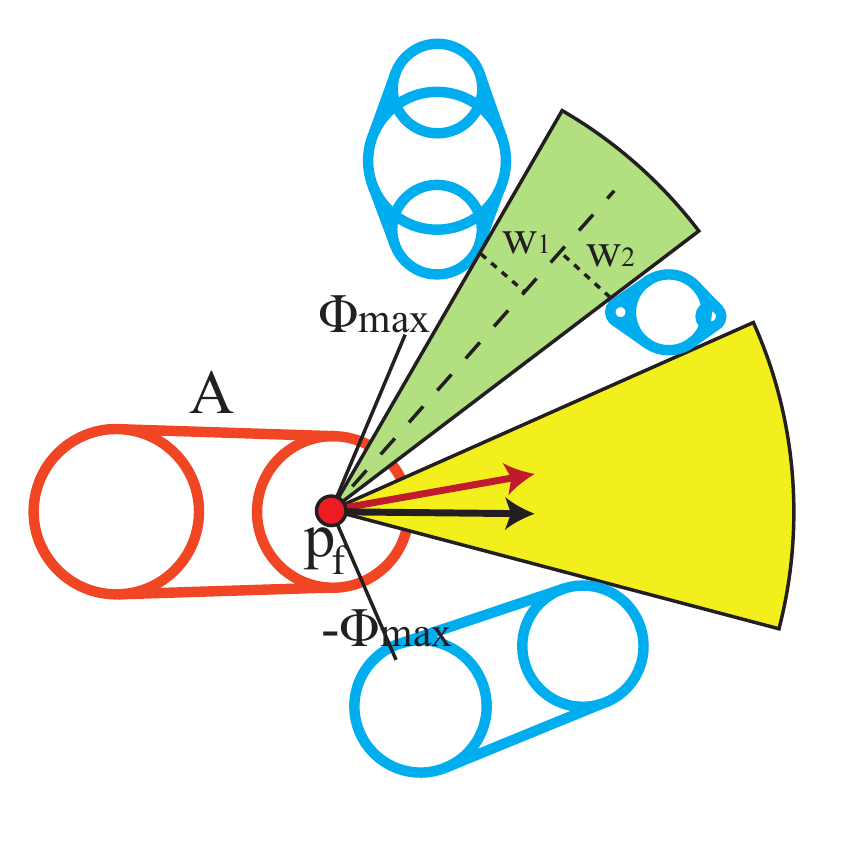}}
\subfigure[]{
\label{fig:searchSpace_2}
\includegraphics[width=0.48\columnwidth]{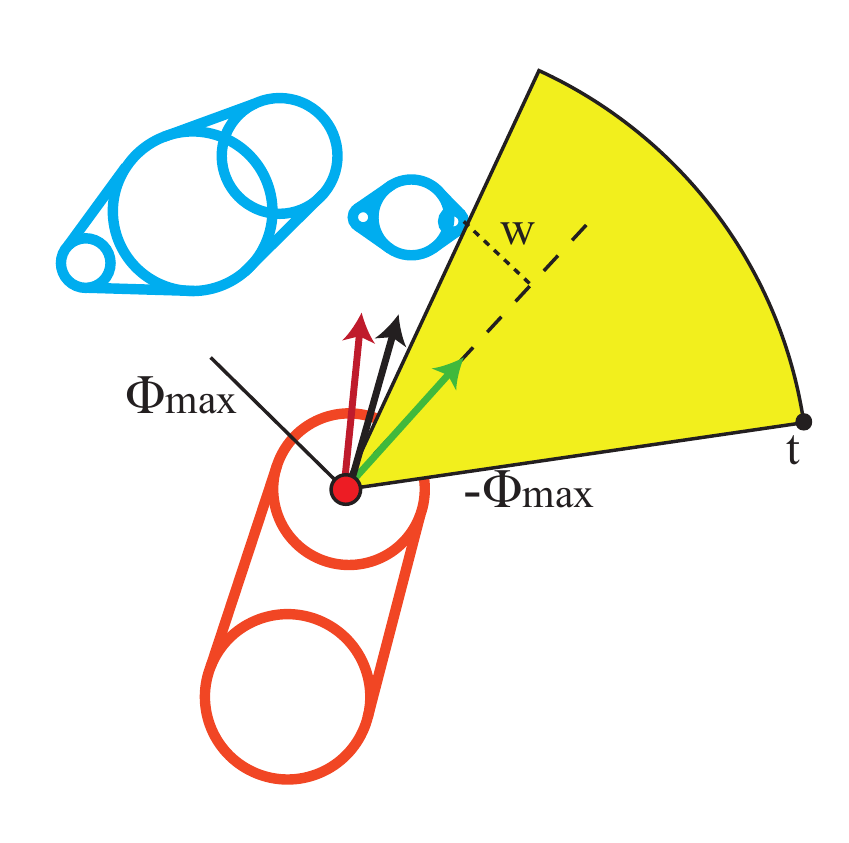}}
\caption{\textbf{Search for free-space for collision-free local navigation}. The black arrow represents the orientation direction $\vec{a}$, the red arrow is the destination direction $\vec{h}$, and the green arrow is the preferred direction $\vec{d}^o$. $\vec{d}^o$ coincides with $\vec{h}$ in (a). All the detected fan spaces for the road-agent in the 2D plane are drawn in green and yellow. The yellow fan denotes the free-space.}
\label{fig:searchSpace}
\vspace{-2ex}
\end{figure}

\subsection{Preferred Velocity Computation}

The trajectory is determined by the velocity of the road-agent. The velocity depends on the speed and steering for vehicles, and hinges on the speed and orientation for pedestrians. Before choosing the velocity for a road-agent, we compute parameters $(v^o, \phi^o)$ or $(v^o, \theta^o)$ first to guide the final velocity selection.

\subsubsection{Preferred Steering Computation}
First, we compute preferred steering $\phi^o$, which is the preferred angle for turning left or right. Let $\vec{a}$ denote the orientation direction of the road-agent. For pedestrians, $\vec{a}$ is the forward-facing direction. For vehicles, $\vec{a} = p_f - p_r$. We define the direction to the destination as $\vec{h}$ and the preferred direction as $\vec{d}^o$. In general, $\vec{d}^o = \vec{h}$. However, in dense traffic scenarios, it is sometimes a good strategy to find a detour space.

 The range of the steering ($-\phi_{max}$, $\phi_{max}$), the considering distance for neighbors and the set of neighbors of the agent are used to compute a set of fan spaces for a vehicle $A$ (Fig.~\ref{fig:searchSpace}). These fan spaces do not contain any obstacle or other road-agent in the detection range. The width of the fan space can be computed by two tangent points of the neighbors and their distance to the middle line of the space. For the green fan space in Fig.~\ref{fig:searchSpace_1}, the width is the sum of $w_1$ and $w_2$. If the side of a space is decided by $-\phi_{max}$ or $\phi_{max}$, like the yellow fan in Fig.~\ref{fig:searchSpace_2}, the tangent point can be replaced by the vertex $t$ of the fan. We regard a fan space as a free-space where is uncrowded, if its width is $\sigma$ times bigger than the road-agent's width ($\sigma=1.5$ in our benchmarks). If there is no feasible space or $\vec{h}$ is already in a free-space (Fig.~\ref{fig:searchSpace_1}), we set $\vec{d}^o = \vec{h}$. Otherwise, $\vec{d}^o$ changes to the green direction (see Fig.~\ref{fig:searchSpace_2}), which satisfies the requirement that $w$ is one half of the width of $A$. Next, according to the angle between $\vec{a}$ and $\vec{d}^o$, we can compute $\phi^o$ for vehicles by a dynamic formulation. 
 \begin{equation}
   \phi^o = f(\vec{a}, \vec{d}^o), \phi^o \in (-\phi_{max}, \phi_{max})
  \end{equation}
 $f$ differs for different types of vehicles and could be computed using real-world data. In terms of pedestrians, they will change their orientation to $\vec{d}^o$ directly.

 \subsubsection{Preferred Speed Computation}
When $\phi \neq 0$, vehicles will move along a circle $C$ with radius $r$ and center $c$ (see Fig.~\ref{fig:model}). According to the centripetal force equation, we can compute the upper bound of the speed:
 \begin{equation}
   v_{{max}_1} = \sqrt{g\mu r}
  \end{equation}
  where $g$ is the acceleration due to gravity and $\mu$ is the friction force coefficient.
  If we take the minimum distance between the road-agent and its neighbors in its current moving direction as $l$. According to vehicle's braking control, we can compute another upper bound of speed:
   \begin{equation}
   l = l_{react} + l_{braking} = v_{{max}_2}t + \frac{{v_{{max}_2}}^2}{2g\mu}
  \end{equation}
where $l_{react}$ is the perception-reaction distance, $l_{braking}$ indicates the braking distance, and $t$ is the time for increasing braking force ($t=1.5$, $\mu=0.7$ for common baseline value). $l$ reflects the minimum safe distance for a specified speed under the dynamic constraint of the braking system of the vehicle. Moreover, each vehicle has its own maximum speed $v_{{max}_3}$.
Therefore, we can get the maximum speed for the road-agent under a specified steering $\phi$: 
  \begin{equation}
v_{max}(\phi) = min\{v_{{max}_1},v_{{max}_2},v_{{max}_3} \}
  \end{equation}
  For pedestrians, we only need to consider $v_{{max}_2}$ and $v_{{max}_3}$. We choose $v^o = v_{max}/2$ in our benchmarks.
  
  \subsubsection{Velocity Prediction}
  
 In the real world, a driver or a pedestrian always has the ability to predict the state and consider that prediction before making further decisions. We therefore compute the state space of the neighboring agents after a time interval $\kappa$, and then compute the free-spaces for the road-agent once again. If the road-agent's $\vec{h}$ was not in a free-space, but after time $\kappa$, $\vec{h}$ is in a free-space, it will choose to stop moving in the next state update, because waiting in such a situation will help the road-agent avoid unnecessary detouring behavior and save energy. If the road-agent's $\vec{h}$ was in a free-space, but after time $\kappa$, $\vec{h}$ is no longer in a free-space, the road-agent will speed up within a reasonable range, because it should pass the uncrowded space as soon as possible to avoid being locked after time $\kappa$.

\begin{figure*}[!t]
\subfigure[]{
\label{fig:simulation_1}
\includegraphics[width=0.5\columnwidth]{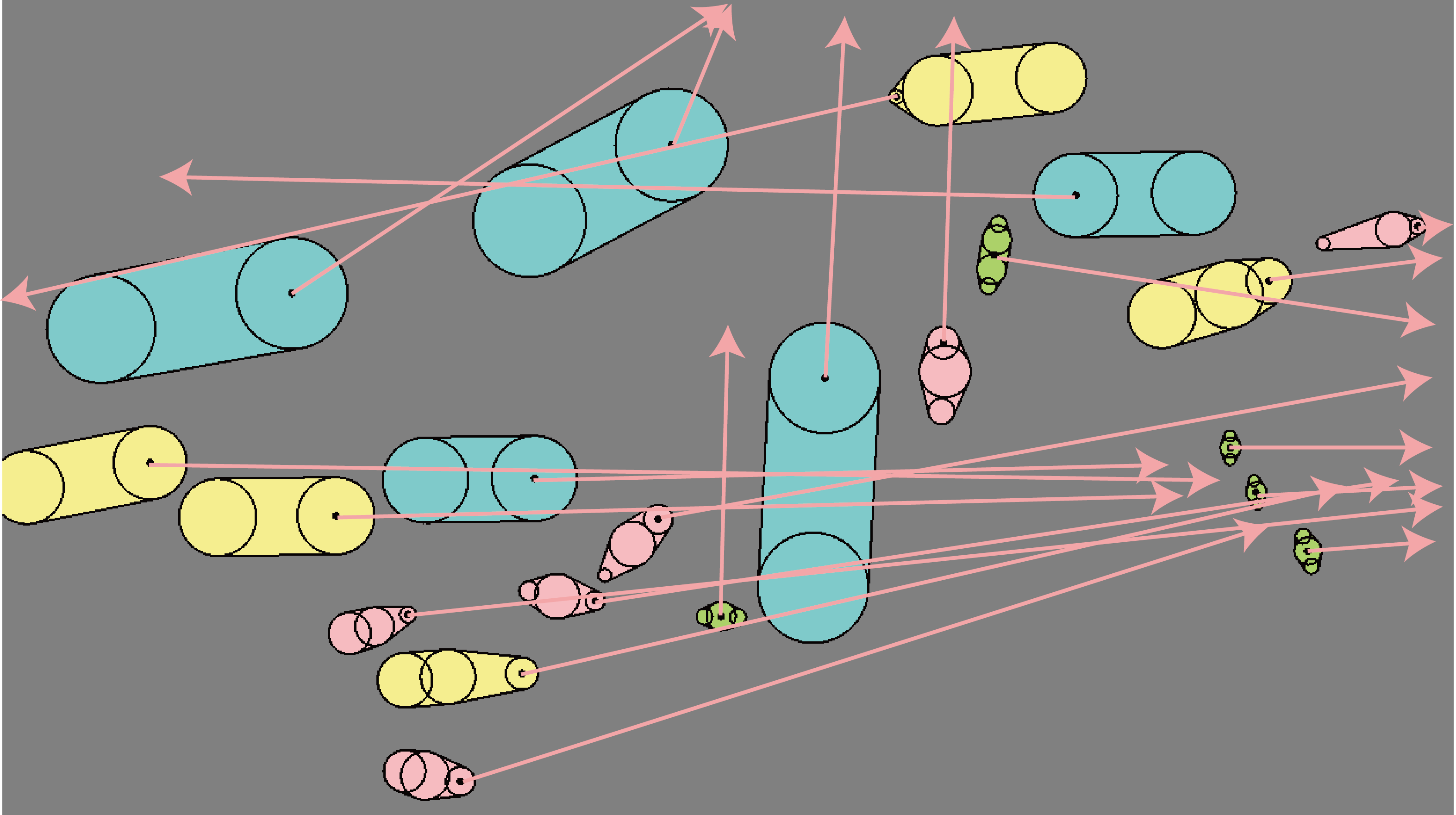}}
\subfigure[]{
\label{fig:simulation_2}
\includegraphics[width=0.5\columnwidth]{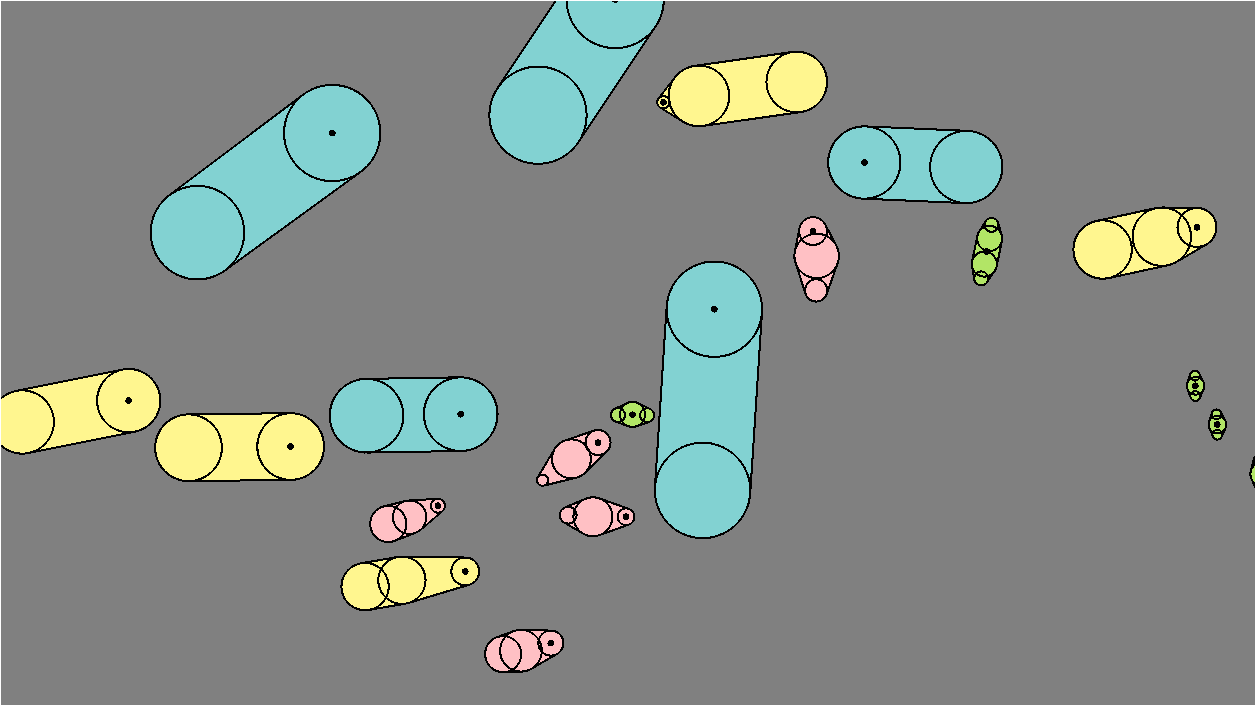}}
\subfigure[]{
\label{fig:simulation_4}
\includegraphics[width=0.5\columnwidth]{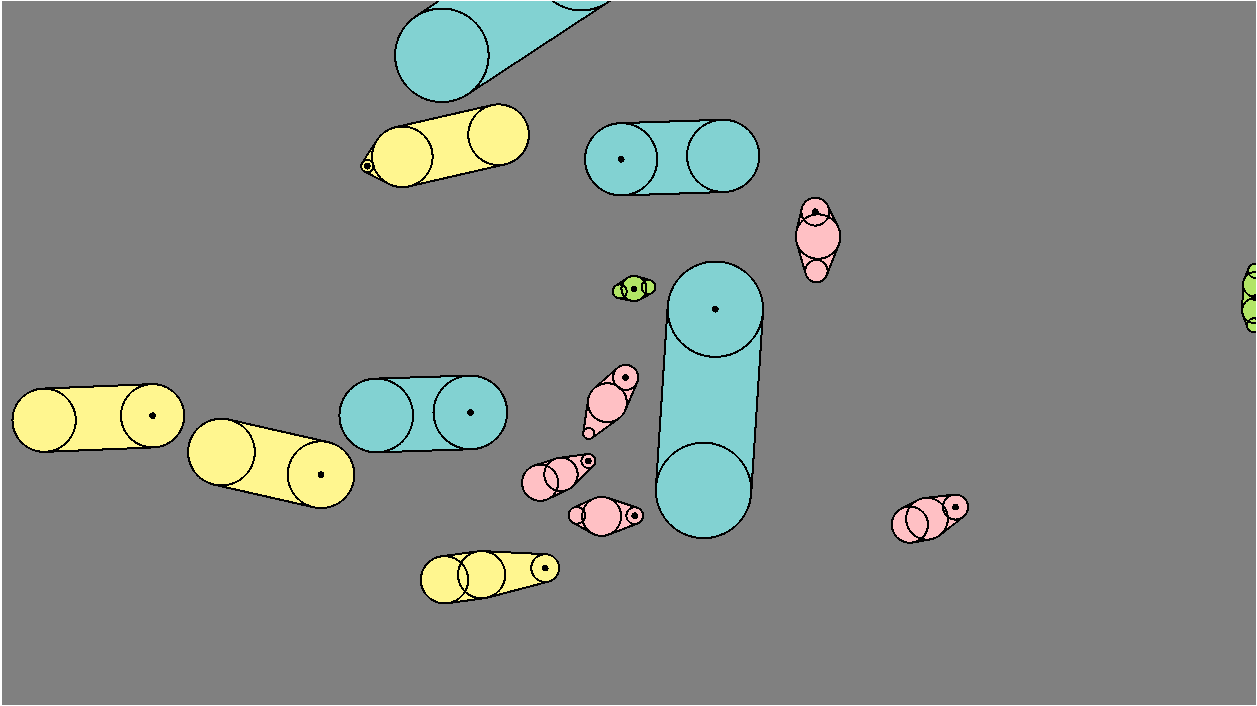}}
\subfigure[]{
\label{fig:simulation_5}
\includegraphics[width=0.5\columnwidth]{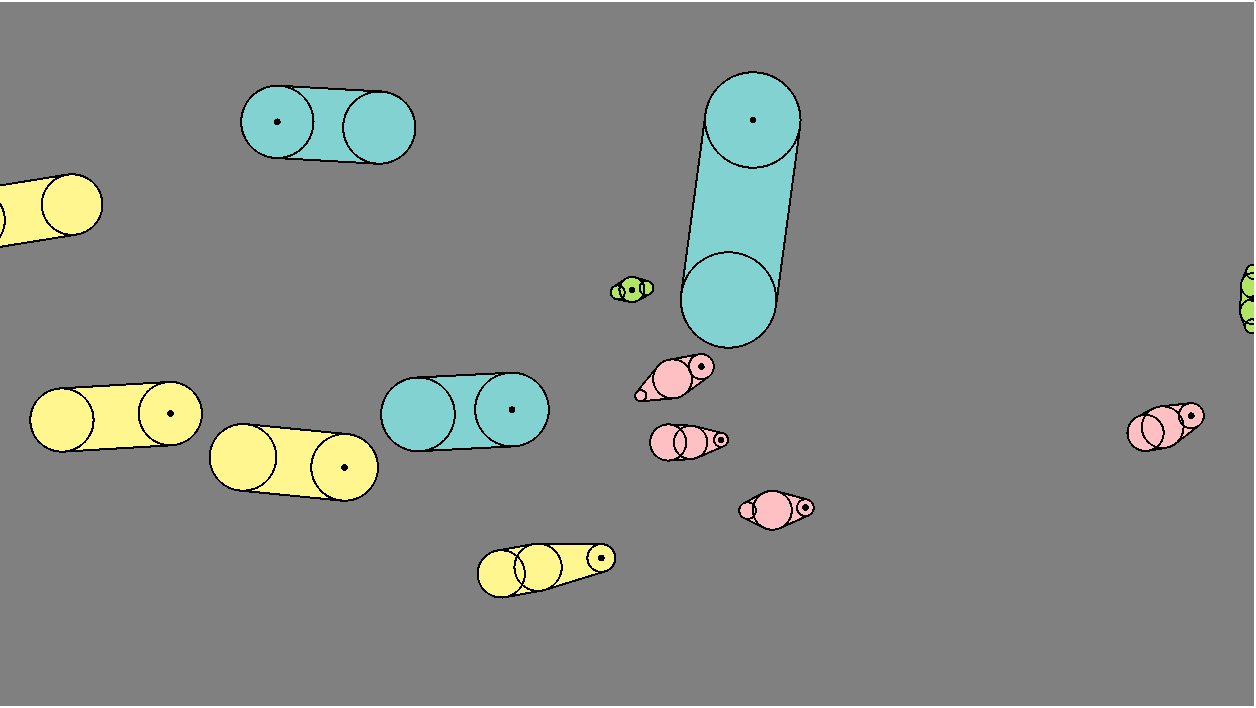}}
\caption{A sequence of frames (selecting frames every 7s) of a simulated dense traffic scenario from the input of Fig.~\ref{fig:traffic}. Red arrows represent designated destinations according to the real video. AutoRVO can compute collision-free trajectories for all road-agents in such dense traffic scenarios.}
\label{fig:result1}
\vspace{-2ex}
\end{figure*}

\subsection{Velocity Sampling}
For disc-based representation, we can use generalized velocity obstacle (GVO)~\cite{bareiss2015generalized} to compute the new collision-free velocity for each vehicle directly. However, such a disk representation can be too conservative. Instead, we use the CTMAT representation, but the resulting computation of the exact boundary of a control obstacle is too expensive. Since we have already know the preferred speed $v^o$ and steering $\phi^o$ of the vehicles, we can use a sampling approach to search for a better solution for $(v, \phi)$. The sampling range can be defined as follows.
\begin{eqnarray}
  (v_{min}, v_{max}) = T(v^o, \tau) \nonumber , \\
  (\phi_{min}, \phi_{max}) = S(\phi^o, \tau) ,
\end{eqnarray}
where $T$ is the dynamic function to get the speed range when the vehicle is using the highest throttle and braking effort for time interval $\tau$. $S$ is the dynamic function to compute the steering range for next $\tau$ time. We perform even sampling in this range. In particular, we choose the collision-free samples as candidates by using the Minkowski sum of CTMAT between the road-agent and its neighbors. We also use Equation 2 to further filter candidates for $(v, \phi)$. We use the same method to sample $(v, \theta)$ for pedestrians.

\subsection{Trajectory Computation}

After computing a set of candidates, we use the following cost function to select the best solution for $(v, \phi)$ or $(v, \theta)$.
   \begin{equation}
   \min F = af_{1}+bf_{2}+cf_{3}+df_{4}+ef_{5} ,
  \end{equation}
  where $a, b, c, d, e$ are coefficients that can be adjusted. They are the weights of making trajectories smoother or safer or faster to arrive the destination.

\begin{equation}
   f_{1} = (v-v^o)^2+(\phi - \phi^o)^2 ,
  \end{equation}
   $f_{1}$ indicates the distance to $v^o$ and $\phi^o$. This term is used to select a solution close to the computed preferred speed and steering. 

\begin{equation}
   f_{2} = \lvert v-v^\prime \rvert+\lvert \phi - \phi^\prime\rvert ,
  \end{equation}
  $f_{2}$ denotes the most recent changes to the previous speed and steering. We use this term to control the changes of vehicles' behaviors and results in smoother trajectories for the vehicles. $(\phi,\phi^o,\phi^\prime)$ are replaced by $(\theta,\theta^o,\theta^\prime)$ for pedestrians in $f_{1}$ and $f_{2}$.
 
\begin{equation}
   f_{3} = -\sum_{n=1}^N{(1+C_{type}-N(n)_{type})\cdot \Vert p - p_n \Vert} ,
  \end{equation}
 where $N(n)_{type}$ is defined as the $n$th neighbor of the road-agent. $p$ is the vehicle's position if current candidate for $(v,\phi)$ or $(v,\theta)$ is adopted. $p_n$ is its $n$th neighbor's position under the assumption that they proceed at their current speed and steering. $f_{3}$ denotes an attempt to keep the vehicle away from nearby pedestrians, vehicles, or obstacles.
  
  \begin{equation}
   f_{4} = -\sum_{n=1}^Nd(S_n, origin) ,
  \end{equation}
We compute whether the coordinate's origin lies in the Minkowski sum of the vehicle and each of its neighbors to determine whether the sample $(v, \phi)$ is collision-free. The probability of causing a collision is lower if the distance from the origin  to the $S$ is bigger. We use $f_{4}$ to reduce the risk of collision.

   \begin{equation}
   f_{5} = d(p, Goal) ,
  \end{equation}
$f_{5}$ stands for the distance to the destination. This term is related to energy consumption. It is better to reach the destination as soon as possible to save the passenger time and to save the vehicle electricity or fuel.

\section{Results}

\begin{figure*}[!t]
\subfigure[traffic-1]{
\label{fig:pathCompare_1}
\includegraphics[width=0.68\columnwidth]{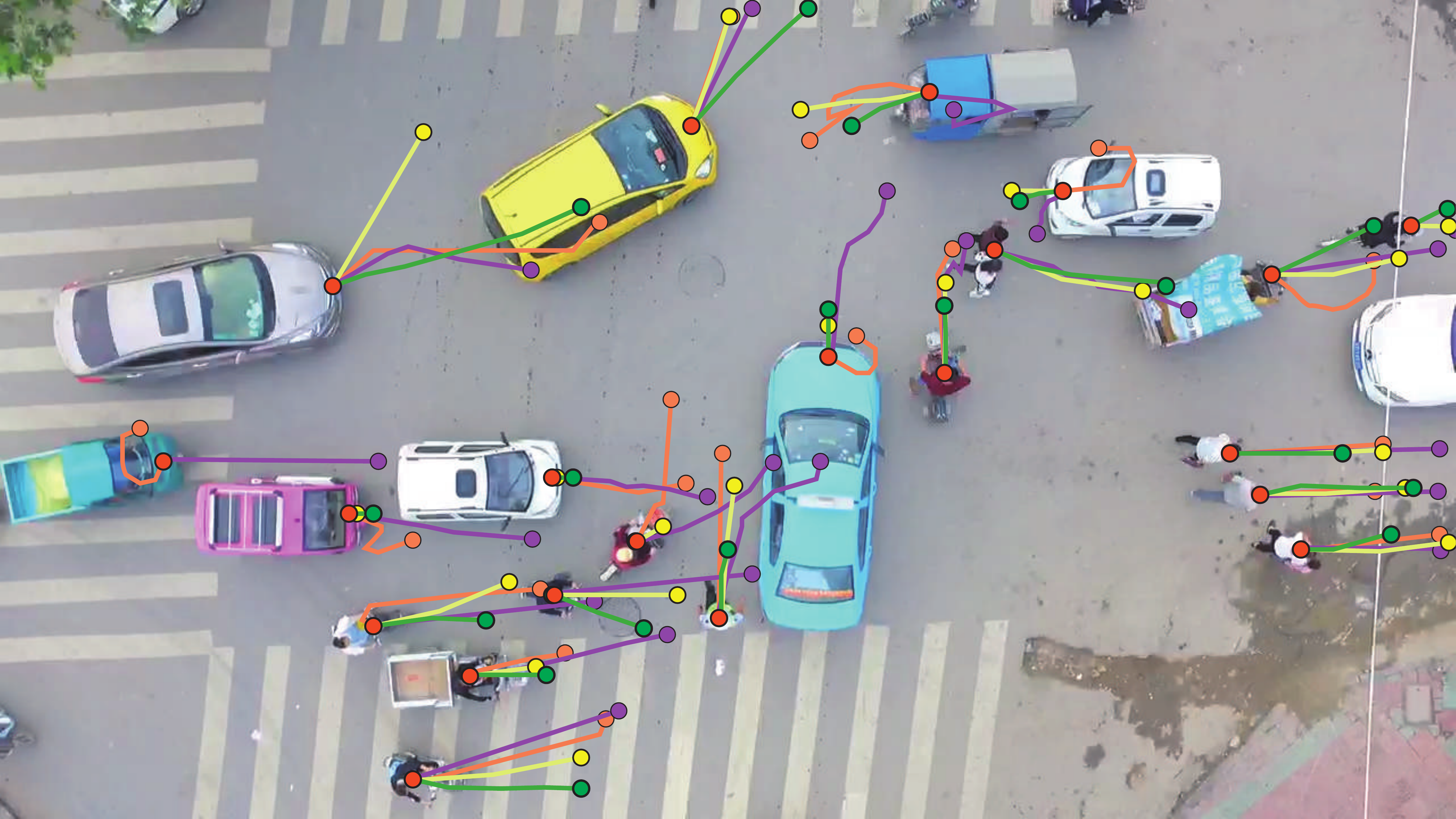}}
\subfigure[traffic-2]{
\label{fig:pathCompare_2}
\includegraphics[width=0.68\columnwidth]{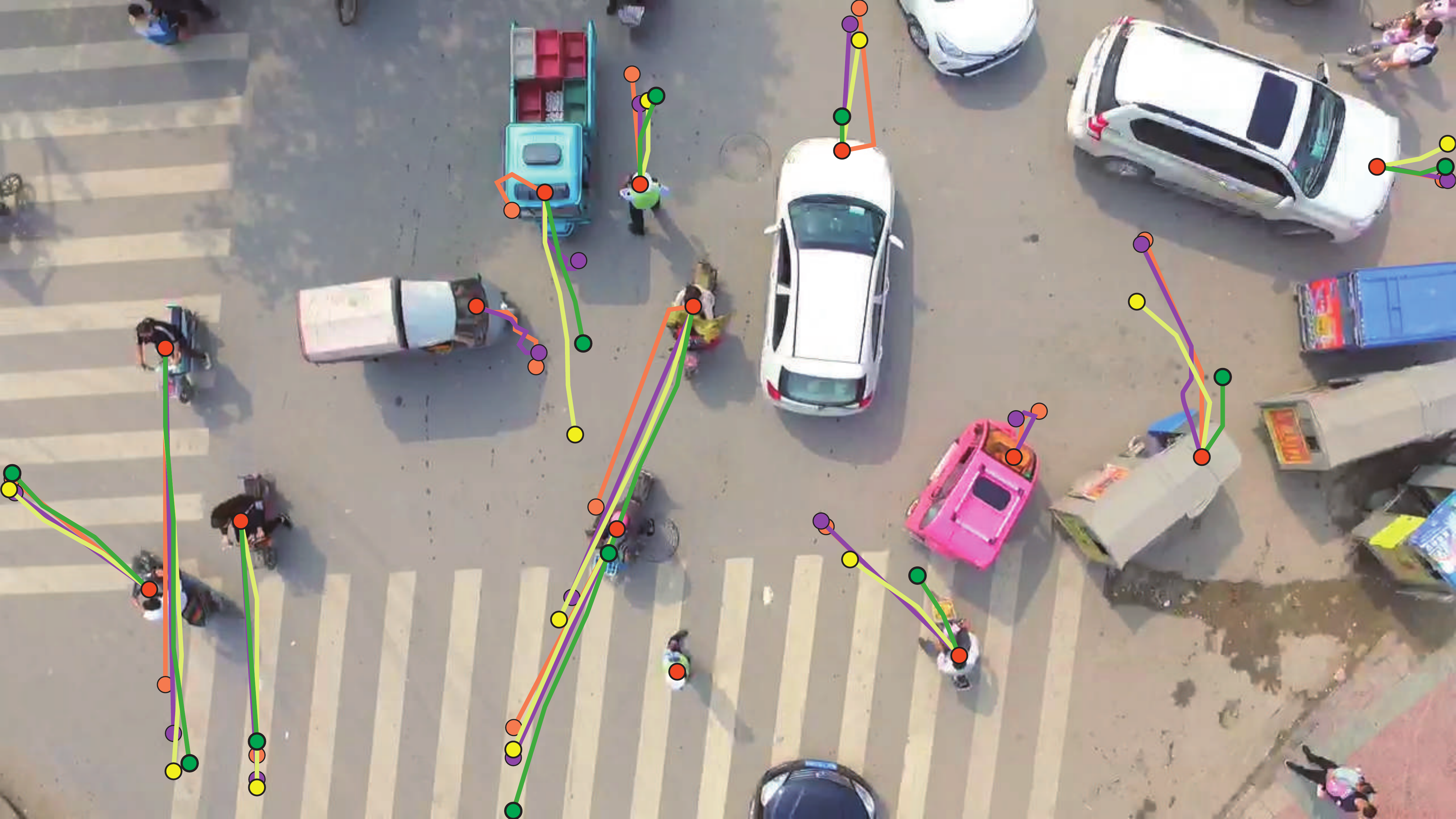}}
\subfigure[traffic-3]{
\label{fig:pathCompare_3}
\includegraphics[width=0.68\columnwidth]{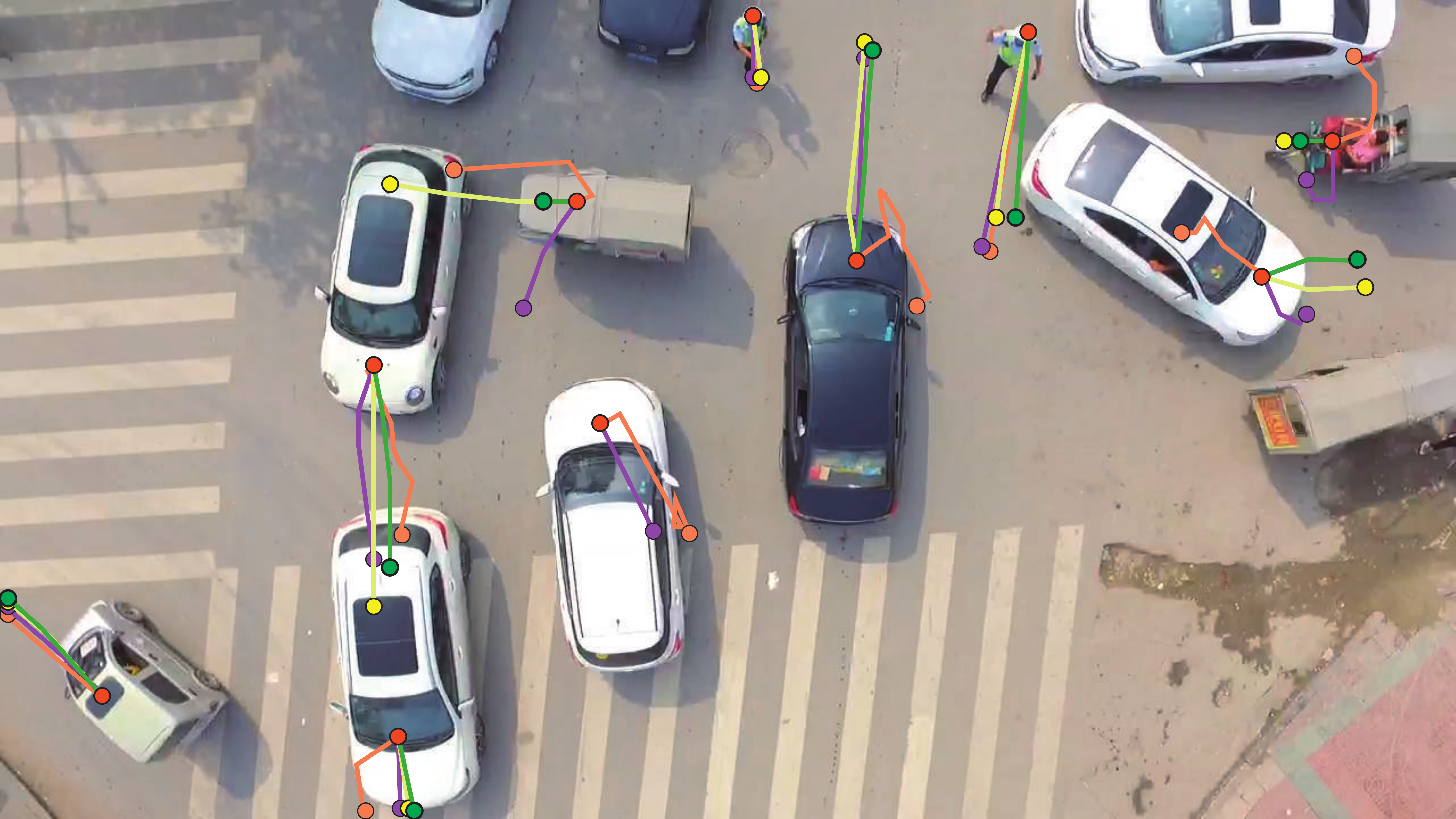}}
\subfigure[traffic-4]{
\label{fig:pathCompare_5}
\includegraphics[width=0.68\columnwidth]{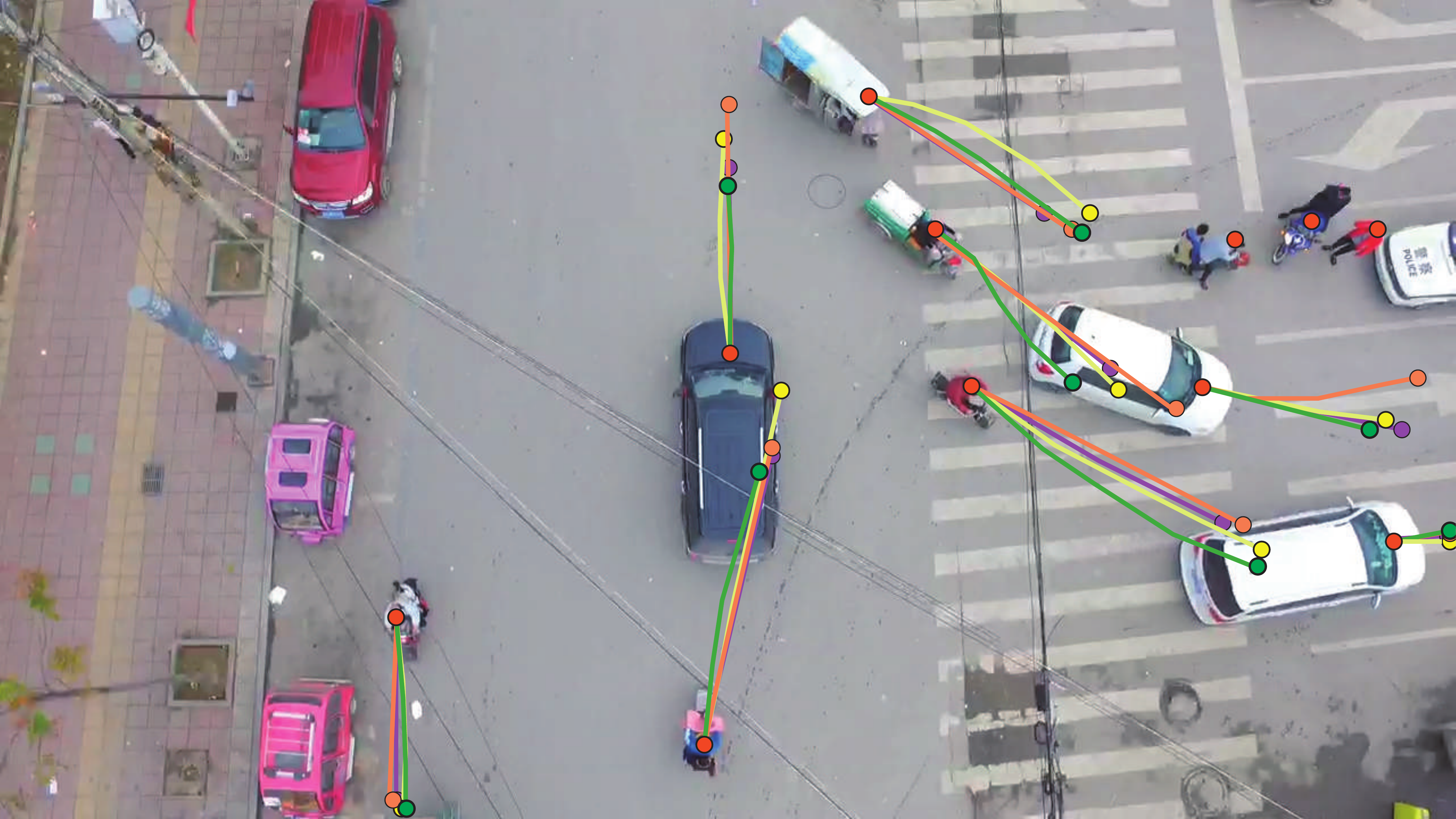}}
\subfigure[traffic-5]{
\label{fig:pathCompare_4}
\includegraphics[width=0.68\columnwidth]{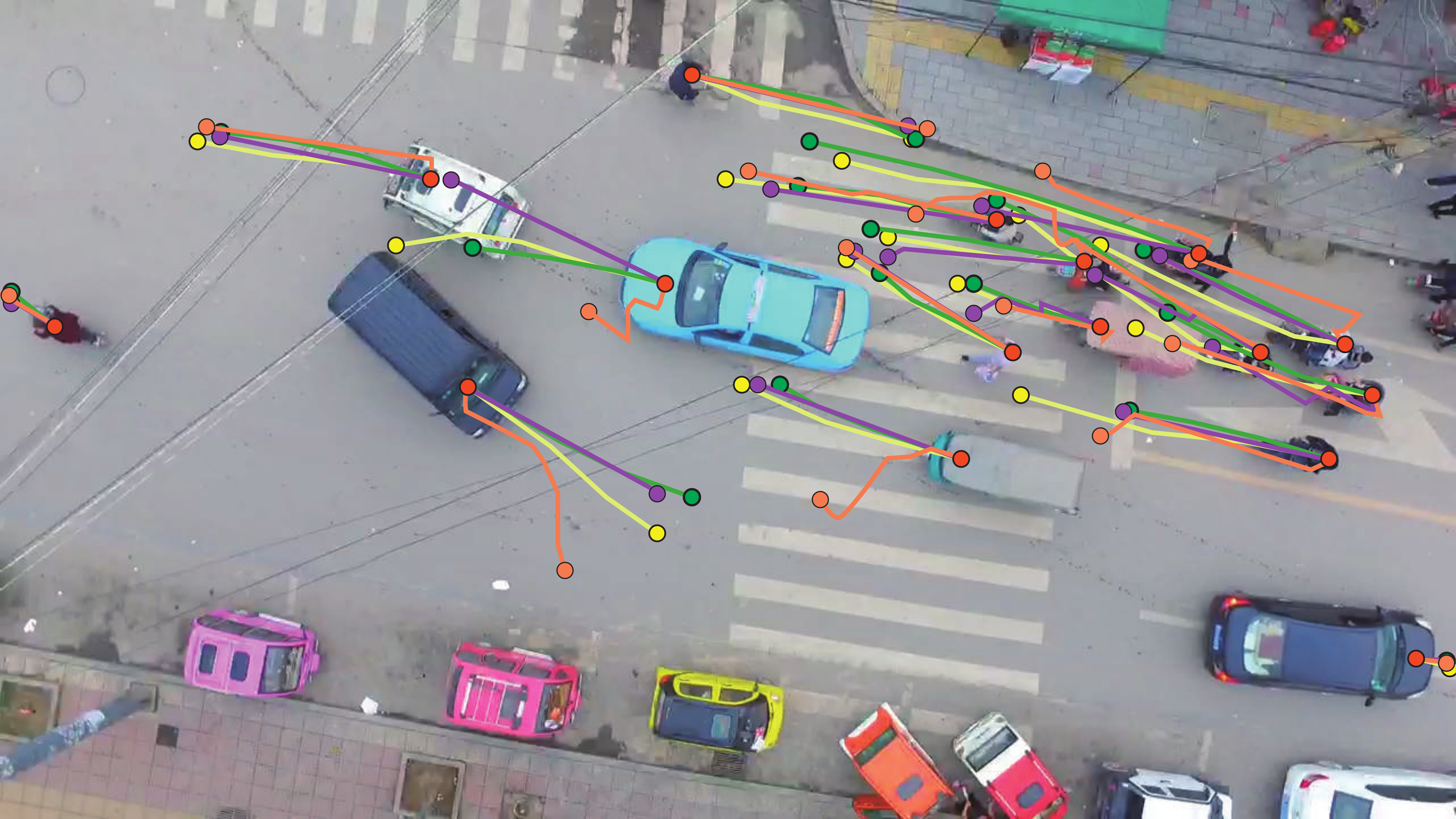}}
\subfigure[traffic-6]{
\label{fig:pathCompare_6}
\includegraphics[width=0.67\columnwidth]{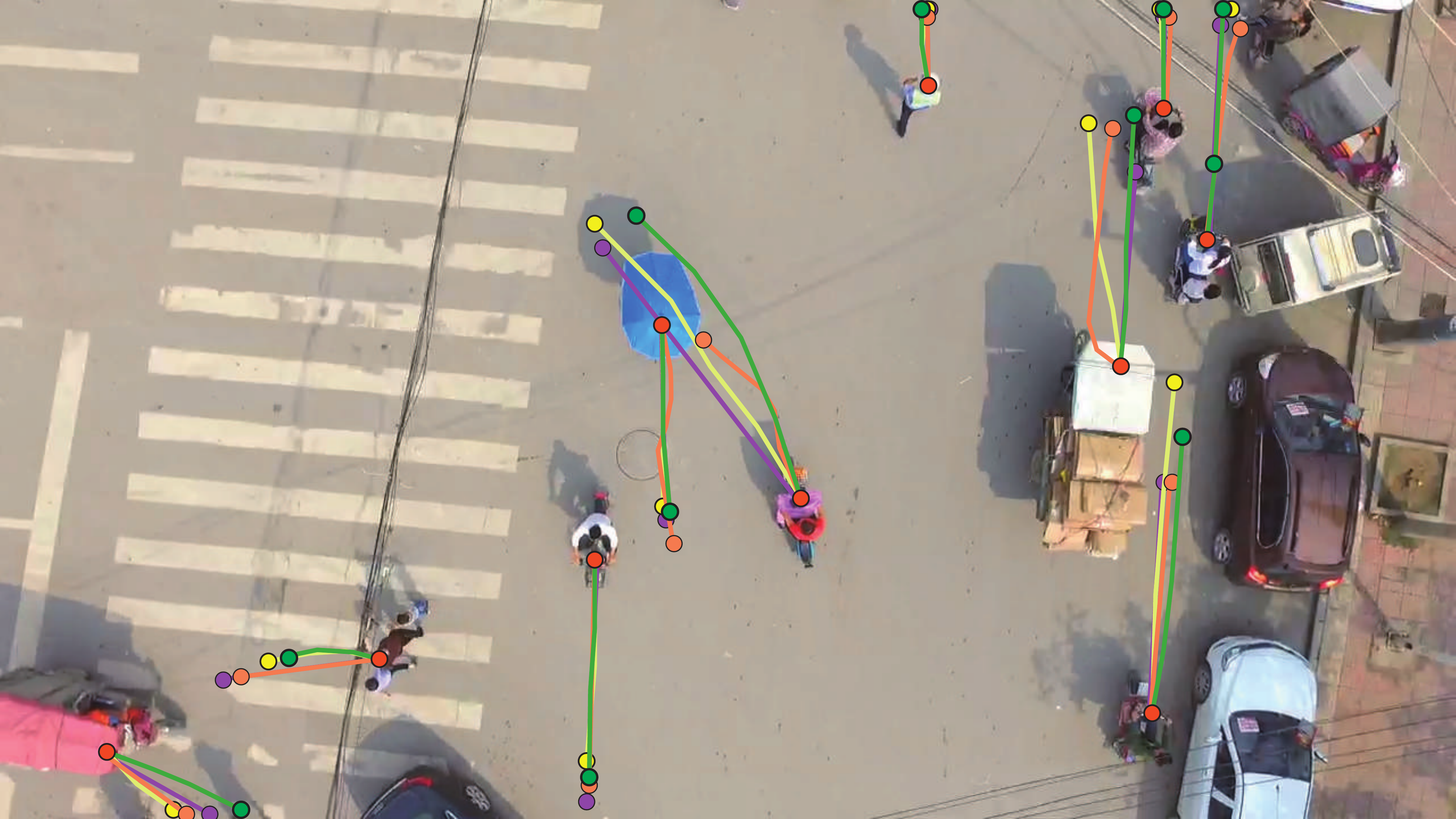}}
\caption{Comparison of real trajectories of 50 continuous frames and simulated trajectories. (a)-(c) are three different moments from one video. (d)-(f) are three moments from three different videos. Green lines indicate the real trajectories extracted from videos captured using a drone. Trajectories generated by AutoRVO, CTMAT representation with no dynamics (CND), and ORCA with disk representation are drawn in yellow, purple, and orange respectively. We observe higher accuracy with AutoRVO. Red points represent beginning reference positions. }
\label{fig:pathCompare}
\vspace{-2ex}
\end{figure*}

In this section, we highlight the performance of our algorithm in local navigation with dynamic constraints in dense scenarios with heterogeneous vehicles. All the traffic scenarios are from a city traffic scene and the original traffic images (Fig. 1 and 6) were captured using a drone camera. The frame rate of all the captured videos is 30fps.
 
Fig.~\ref{fig:result1} shows a sequence of frames in our simulation of dense traffic with different vehicles. Before we run our algorithm, we select any frame in one video as the input and then compute CTMAT representation according to the road-agents' contours (see Fig.~\ref{fig:traffic}). Because the view for a given camera is limited, we assign goal positions for road-agents based on the corresponding positions where they stop or disappear in the video. We represent the destinations of vehicles and pedestrians using the red arrows in Fig.~\ref{fig:simulation_1}. These traffic scenarios are very dense and include various types of vehicles and pedestrians. As we can see from the navigation results generated by our algorithm, all the pedestrians and vehicles move in collision-free trajectories and behave realistically waiting or detouring behaviors without creating any gridlocks or congestion scenarios. 

 Fig.~\ref{fig:pathCompare} shows comparisons between road-agents' trajectories and simulated trajectories of AutoRVO, CTMAT representation without dynamic constraints (denoted as CND) and ORCA. We use 50 continuous frames of a video as one sample to make the comparisons. For each sample, we take the first frame as input and use the positions of road-agents when they disappear or their positions after $100$ frames as the destinations. Then, we select similar number of frames or discrete positions of simulated video for comparison. We can see from the simulation results by AutoRVO that, apart from exhibiting similar trajectories, some road-agents wait during this period as in real-world scenarios, which means our prediction ability also works in solving congestion. In Table 1, we use Entropy metric~\cite{guy2012statistical} to measure the similarity between simulated trajectories by three algorithms and real trajectories. The Entropy metric compares the accuracy of the trajectories computed by different simulated algorithm with real-world trajectories extracted from videos. A lower value of Entropy metric indicates higher accuracy (as observed for AutoRVO). By adding dynamic constraints, we observe considerable accuracy improvement in AutoRVO over CND. The disk representation in ORCA-based methods cannot compute accurate trajectories in such cases. Especially for very dense scenarios like traffic-1 and traffic-5, the Entropy Metric values of CND and ORCA are much bigger than that of AutoRVO, which illustrates our capability in handling dense traffic situations. CND performs slightly better than AutoRVO in traffic-4. This is due to the fact that we use some cost functions in Equations (6)-(11) to compute smoother and safer trajectories. In this case, these functions result in trajectories different from the real-world. But in most cases, our algorithm performs much better than others.

\section{Runtime Analysis}
\begin{figure}[!t]
\subfigure[]{
\label{fig:simulation_1}
\includegraphics[width=1\columnwidth]{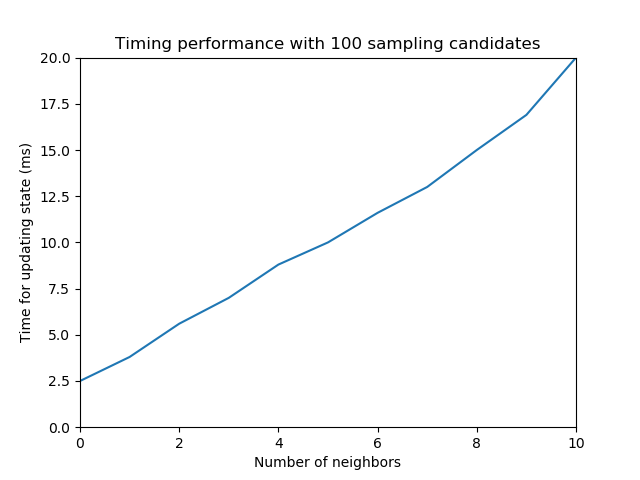}}
\subfigure[]{
\label{fig:simulation_2}
\includegraphics[width=1\columnwidth]{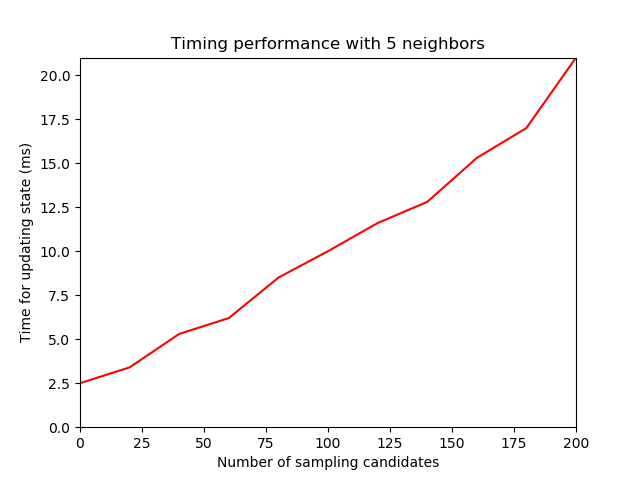}}
\caption{Timing performance on updating the state for one traffic-agent. (a) Timing performance with 100 sampling candidates. (b) Timing performance with 5 neighboring traffic agents.}
\label{fig:timePerformance}
\vspace{-2ex}
\end{figure}

We implemented the algorithm in C++ and conducted experiments on a Windows 10 laptop with an Intel i7-6700 CPU and 8GB RAM. Our algorithm can be parallelized on multiple cores, but we generate all the results on a single CPU core. For the benchmarks shown in Fig.~\ref{fig:result1}, the average number of neighbors for a vehicle is 4, and the average time for updating the state of one road-agent is about 10ms with about $100$ sampling candidates of $(v, \phi)$. Timing performance on fixed sampling candidates and on fixed neighbors is shown in Fig.~\ref{fig:timePerformance}. With the number of neighbors or the number of samples increase, the simulation time per frame increases linearly. The run time $R$ for updating the state space for road-agent $A$ is given as:
  \begin{equation*}
   R(A) = N\cdot M \cdot t_{sum}(A)+t_{search},
  \end{equation*}
where $N$ is the the number of neighboring obstacles and other road-agents of $A$, $M$ is the number of sampling points of $(v, \phi)$, and $ t_{sum}(A)$ indicates the average time of computing the Minkowski sum of a pair of CTMAT. $t_{search}$ is $O(N)$ for searching detour space. Therefore, $R$ is $O(NM)$. 

\begin{table}
\begin{tabular}{|c|c|c|c|c|c|c|}
\hline
Scenario &V&P &T & AutoRVO & CND & ORCA \\
\hline
traffic-1 & 16 &5 &4 & \bf{3.77} & 12.72  &15.11 \\
\hline
traffic-2 & 12& 2 &4 & \bf{2.56} & 5.34 & 8.12  \\
\hline
traffic-3 & 8 & 2 &3 & \bf{2.69} & 8.98 & 10.13  \\
\hline
traffic-4 & 10 &1 &4 & 4.25 & \bf{4.03} & 5.41\\
\hline
traffic-5 & 15 &1 &4 & \bf{2.45} &  5.77& 14.12  \\
\hline
traffic-6 & 8 &2 &3 & \bf{3.33} & 4.33 & 5.63  \\
\hline
\end{tabular}
\label{table:validation}
\caption{Evaluation of different multi-agent navigation algorithms on dense traffic scenarios shown in Fig.~\ref{fig:pathCompare}. We show the total number of vehicles in the second column and the number of pedestrians in the third column. The number of different types of road-agents is shown in the third column. The last three columns illustrate the Entropy metric (lower is better for validation) for different simulation results by our algorithm AutoRVO, CTMAT representation with no dynamics (CND), and ORCA algorithm with disk representation.}
\vspace{-2ex}
\end{table}

\section{Conclusion and limitations}
We present a novel algorithm, AutoRVO, for local navigation of heterogeneous vehicles and pedestrians in dense traffic situations with kinematic and dynamic constraints. Our formulation is based on a tight-fitting media-axis-based agent representation and we present an efficient algorithm to handle kinematic and dynamic constraints of different vehicles and pedestrians. We use an optimization-based local planning method to help road-agents choose a velocity that would result in smooth trajectories.. We have demonstrated the performance of our algorithm in the simulation of dense traffic scenarios and compared its performance with the trajectories of real-world road-agents and other algorithms.

Our approach has some limitations. We assume perfect sensing abilities in terms of the exact position and velocity of all road-agents. In terms of dynamic constraints, we make use of empirical values for some parameters in our equations corresponding to the motion computation, which may different from the real-world data in the videos. In the future, we would like to consider sensor errors and extend our algorithm to handling noisy perception data. Furthermore, we will collect data from different types of vehicles and environmental information to make the dynamic constraints of road-agents closer to real conditions, We will like to combine AutoRVO with data-driven methods to generate plausible traffic scenarios for autonomous driving simulators. 

\section{Acknowledgements}
This research of Dinesh Manocha is supported in part by ARO grant W911NF16-1-0085, and Intel. This research of Wenping Wang is partially supported by the Research Grant Council of Hong Kong (HKU 717813E).